\documentclass[a4paper,fleqn]{cas-dc}

\usepackage[numbers,sort&compress]{natbib}
\usepackage{amsmath,amsfonts,amssymb}
\usepackage{graphicx}
\usepackage{array,tabularx,booktabs, multirow,makecell}
\usepackage{url,textcomp}
\def\tsc#1{\csdef{#1}{\textsc{\lowercase{#1}}\xspace}}
\tsc{WGM}
\tsc{QE}
\tsc{EP}
\tsc{PMS}
\tsc{BEC}
\tsc{DE}

\begin{document}
\let\WriteBookmarks\relax
\setcounter{topnumber}{3}
\setcounter{dbltopnumber}{2}

\renewcommand{\topfraction}{0.95}
\renewcommand{\dbltopfraction}{0.95}
\renewcommand{\textfraction}{0.05}
\shorttitle{TSGPD-IR for All-in-One Infrared Restoration}
\shortauthors{Xinyao Wang et~al.}

\title [mode = title]{Breaking Weather-Content Coupling: Type-Severity Guided Progressive Disentanglement for All-in-One Infrared Restoration}                      
\tnotetext[1]{This work is supported by the National Natural Science Foundation of China under Grant No. 62531012, the Sichuan Science and Technology Program under Grant 2026YFHZ0205, and the XJTU Research Fund for AI Science, No.2025YXYC004.}
\author{Xinyao Wang}[orcid=0009-0009-4995-4642]
\ead{wangxy04@stu.xjtu.edu.cn}
\credit{Writing – review, editing, Writing – original draft,
Methodology, Data curation, Conceptualization}

\affiliation{organization={Shaanxi Key Laboratory of Deep Space Exploration Intelligent Information Technology, School of Information and Communications Engineering, Xi'an Jiaotong University},
                city={Xi'an},
                postcode={710049},
                state={Shaanxi},
                country={China}}

\author{Lijun He}
\cormark[1]
\ead{lijunhe@mail.xjtu.edu.cn}
\credit{Writing – review, editing, Methodology, Conceptualization}

\author{Zhihan Ren}
\ead{renzh@stu.xjtu.edu.cn}
\credit{Methodology, Conceptualization, Validation}

\author{Fan Li}
\ead{lifan@mail.xjtu.edu.cn}
\credit{Methodology, Conceptualization}

\cortext[cor1]{Corresponding author}

\begin{abstract}
Infrared (IR) imaging is crucial for autonomous driving, remote sensing, and other perception tasks. However, adverse weather may introduce fake structural responses that are entangled with real thermal structures. Existing IR restoration methods are typically designed for a single degradation type or directly reconstruct from degradation-entangled representations. Consequently, they struggle to distinguish intrinsic thermal structures from weather-induced fake responses and to accommodate spatially varying degradation severity, leading to artifacts or the over-suppression of weak but meaningful thermal responses. To address these issues, we propose TSGPD-IR, a type-severity guided progressive disentanglement network for all-in-one infrared restoration that factorizes restoration guidance into task-level weather semantics and region-level degradation severity. Specifically, a Weather and Semantic Co-Guided Multi-Level Prompt Generation Module combines global weather semantics with stage-wise local features to generate adaptive prompts that progressively suppress degradation-induced responses while preserving intrinsic thermal structures. To complement global weather semantics with spatial restoration control, a Proxy-Supervised Regional Degradation Estimator derives severity supervision without manual annotations and predicts spatially varying degradation priors. Guided by these cues, a Multi-Source Collaborative Expert Selection Strategy uses a shared branch to preserve weather-invariant thermal structures and hierarchical routing to select weather-specific expert pools and severity-compatible regional experts. This design progressively separates degradation interference from genuine thermal content and enables region-adaptive restoration, reducing both residual artifacts and over-suppression. Experiments demonstrate that TSGPD-IR achieves competitive performance on both all-in-one multi-weather and single-weather infrared restoration tasks.
\end{abstract}



\maketitle

\section{Introduction}
With the development of infrared sensing and deep learning, infrared (IR) imaging has become increasingly important for autonomous driving, object tracking, and urban scene perception. Unlike visible-light cameras, which rely on reflected light, IR cameras sense thermal radiation and can operate at night and in low-light conditions. However, adverse weather can reduce thermal contrast, blur target boundaries, and introduce spatially nonuniform artifacts and structural distortions. These degradations impair downstream perception tasks \cite{ren2023context,he2024unsupervised}, motivating the need for robust adverse-weather IR restoration.

Adverse-weather IR restoration is challenging because weather-induced artifacts are closely coupled with intrinsic thermal structures. Weak target contours and low-contrast background responses can be confused with weather-induced attenuation or occlusion, making it difficult to remove artifacts without suppressing meaningful thermal information. Moreover, degradation severity may vary substantially across regions within a single image with variations in imaging distance, viewing geometry, and local weather interference. Mildly degraded regions may retain reliable thermal responses, whereas severely degraded regions require stronger restoration. Such spatial heterogeneity therefore calls for region-adaptive restoration rather than a fixed restoration strength.

Early adverse-weather restoration methods were generally designed for a single degradation type, such as haze, rain, or snow \cite{2018desnownet}. Although these degradation-specific methods can perform well under predefined conditions, they generalize poorly across weather types. To handle multiple degradation types with a single model, all-in-one visible-image restoration methods employ degradation representations \cite{2022all}, learnable weather embeddings \cite{2022transweather}, weather-general and degradation prompts \cite{2023promptir}. Recent methods have also incorporated degradation severity, spatial priors, and adaptive routing \cite{utilityir, ren2024super}. However, these methods are primarily developed for visible imagery, where degradation cues differ from thermal responses in IR images. Consequently, weather-induced interference can be confused with weak intrinsic thermal variations, resulting in residual artifacts or the suppression of meaningful thermal structures.

IR restoration has evolved from joint tone mapping and denoising to thermal dehazing, compound-degradation enhancement \cite{2026enhancing}, and joint correction of blur, rolling-shutter distortion, and noise. More recently, unified IR methods have been developed to address noise, blur, low contrast, and their predefined combinations \cite{2026breaking,2026expandable}. However, these methods mainly target sensor or imaging-related degradations and their predefined combinations, whereas adverse-weather degradation is often spatially nonuniform and closely coupled with scene content. Applying a common restoration process to such heterogeneous degradation can lead to inappropriate restoration strengths across regions, causing over-suppression in mildly degraded areas and residual artifacts in severely degraded ones.

In summary, existing methods face two key problems: (1) \textbf{Strong coupling between weather degradation and intrinsic thermal structures:} Weather artifacts, target contours, and background thermal distributions are often entangled in the feature space, making it difficult to suppress degradation without damaging weak but meaningful thermal details. (2) \textbf{Limited region-adaptive restoration capability:} Because weather degradation in IR images is spatially nonuniform, applying a uniform restoration strength may over-suppress mildly degraded regions while leaving residual artifacts in severely degraded ones. These challenges motivate joint modeling of global weather type and regional severity to determine the restoration direction and local processing strength.

To address these problems, we propose TSGPD-IR, a Type-Severity Guided Progressive Disentanglement Network for all-in-one adverse-weather IR restoration. Rather than directly mapping degradation-entangled representations to clean outputs, TSGPD-IR factorizes restoration guidance into global weather type and regional degradation severity, which respectively determine the weather-specific restoration direction and region-wise restoration strength. Based on this design, the Weather and Semantic Co-Guided Multi-Level Prompt Generation Module (WS-MPG) provides stage-adaptive guidance, the Proxy-Supervised Regional Degradation Estimator (PS-RDE) estimates regional degradation severity, and the Multi-Source Collaborative Expert Selection Strategy (MCESS) performs hierarchical type-severity routing. Their collaboration progressively suppresses weather interference while preserving meaningful thermal structures.

The main contributions are summarized as follows:
\begin{itemize}
\item[1)] \textbf{Type-Severity Guided Disentanglement of Degradation Interference and Thermal Structures.}
Weather-induced interference and intrinsic thermal content are strongly coupled in degraded IR features. Existing methods generally perform restoration directly on such entangled representations without explicitly separating the two, making the restoration process susceptible to degradation interference and leading to residual artifacts or the suppression of weak thermal structures. To address this problem, we propose TSGPD-IR, a progressive type-to-severity disentanglement framework that separates weather-induced spurious responses from intrinsic thermal structures, thereby suppressing degradation interference while preserving genuine thermal content.

\item[2)] \textbf{Multi-Source Region-Adaptive Restoration for Spatially Nonuniform Degradation.}
To model spatial variations in degradation severity and the resulting region-dependent restoration requirements, WS-MPG injects multi-level weather-semantic priors at different restoration stages to progressively refine degraded representations, while PS-RDE produces region-level severity cues to distinguish heavily degraded regions from mildly affected ones. MCESS further preserves weather-invariant thermal structures through a shared branch, selects a weather-specific expert pool, and then routes different regions to severity-compatible experts. This multi-source collaboration allocates restoration capacity according to local degradation conditions, reducing under-restoration in severely degraded regions and over-restoration in mildly affected ones while preserving reliable thermal structures.
\end{itemize}

\section{Related Work}
\subsection{All-in-One Weather Visible Image Restoration}

Early adverse-weather restoration methods typically targeted a single degradation type using dedicated physical models or visual priors. Representative examples include the dark channel prior for image dehazing and explicit snow-particle modeling for image desnowing \cite{2018desnownet}. Although effective under predefined conditions, these methods generally require separate models for different weather types and are difficult to apply when the degradation type is unknown or multiple weather effects coexist.

All-in-one restoration methods alleviate this limitation by handling diverse degradations within a unified framework. TransWeather \cite{2022transweather} introduced weather queries into a Transformer-based encoder-decoder, while PromptIR \cite{2023promptir} further employed learnable degradation prompts to dynamically modulate restoration features. More recent studies have explored explicit degradation modeling and semantic guidance for unified restoration. Ada4DIR combines model-driven degradation modeling with prompt learning to adaptively identify and restore multiple degradation types \cite{2025ada4dir}. Related studies in infrared-visible fusion have further demonstrated the effectiveness of textual semantics for adaptive visual processing. TextFusion introduces textual semantics to control multimodal fusion \cite{2025textfusion}, while InstructIVF employs diverse textual instructions to achieve degradation-aware infrared-visible fusion \cite{2025instructivf}. \citet{2024SDCFusion} further proposed a semantic-driven coupled network for infrared and visible image fusion, demonstrating the effectiveness of semantic information guidance in multimodal image fusion.

These approaches improve restoration adaptability across different weather conditions. However, their degradation representations and restoration mechanisms are primarily developed for visible imagery and do not specifically address the ambiguity between weather interference and weak thermal structures in IR images.
\subsection{Infrared Image Restoration}

Existing IR restoration methods mainly address sensor and imaging-related degradations, including random noise, fixed-pattern noise, low contrast, blur, and limited spatial resolution. These studies indicate that degradation models developed for visible images may not transfer directly to IR imagery because thermal sensors exhibit distinct imaging responses and noise characteristics.

Fewer studies focus on atmospheric or weather-induced IR degradation. InfDiff \cite{2024adapting} addressed rainy-weather infrared restoration by adapting high-quality priors generated by a visible-image diffusion model. DparNet \cite{2025infrared} estimated spatially varying degradation parameters for atmospheric-turbulence correction. Although these methods demonstrate the necessity of IR-specific atmospheric modeling, they generally focus on an individual weather or atmospheric condition.

More recently, unified IR restoration methods have been proposed for multiple or compound degradations. PPFN \cite{2026enhancing} progressively incorporates degradation-related prompts to handle noise, blur, low contrast, and their combinations. ECMRNet \cite{2026expandable} further addresses open-world IR restoration by expanding, compressing, and reusing representations for newly encountered degradations.

Despite this progress, existing all-in-one IR approaches predominantly focus on sensor and imaging-related degradations, whereas weather-oriented IR restoration remains dominated by condition-specific solutions. Unified modeling of multiple adverse-weather types and their spatially varying severities in IR imagery therefore remains unexplored.

\begin{figure*}[h]
\centering
\includegraphics[width=\textwidth]{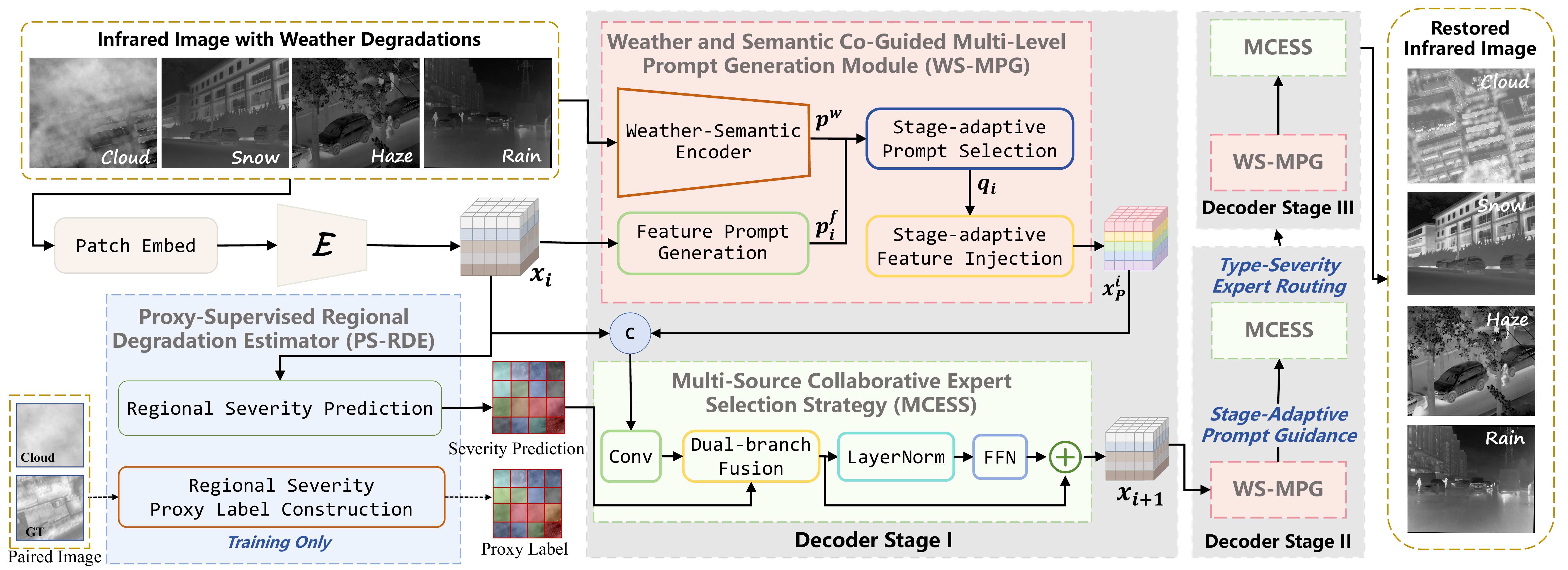}
\caption{Overall pipeline of TSGPD-IR. WS-MPG provides stage-adaptive weather-semantic prompts, PS-RDE predicts spatially varying degradation severity, and MCESS converts the type and severity cues into hierarchical expert routing.}
\label{fig_1}
\end{figure*}

\subsection{Spatially Adaptive Degradation Modeling and Expert Routing}

Spatially adaptive restoration adjusts image processing according to local degradation characteristics. DparNet \cite{2025infrared} learns degradation-parameter matrices for spatially and intensity-adaptive correction. These studies demonstrate the importance of spatially adaptive guidance for heterogeneous degradation.

Degradation-aware restoration has also been combined with severity modeling and expert routing. UtilityIR \cite{utilityir} models degradation type and severity and injects them through local-global adaptive modulation. MoCE-IR \cite{2025complexity} further introduces experts with different computational complexities and receptive fields for all-in-one restoration.

These methods establish foundations for severity-aware modulation, spatial degradation modeling, and adaptive expert routing. TSGPD-IR differs from them by deriving a proxy-supervised spatial severity field from residual and residual-gradient statistics of paired IR images and nesting severity-based expert selection within a weather-specific task pool. Specifically, global weather semantics first select the task-level expert pool, regional severity then selects a severity-compatible expert within the activated pool, and stage-wise prompts modulate the selected expert responses. Thus, TSGPD-IR combines IR-specific regional severity supervision with hierarchical routing from weather pools to severity experts.

\section{Method}
\subsection{Overview of the Proposed Framework}
Under adverse weather conditions, including haze, rain, snow, and cloud occlusion, infrared images often suffer from reduced thermal contrast, blurred details, and distorted local thermal structures. Unlike conventional single-degradation restoration, all-in-one multi-weather infrared image restoration requires disentangling weather-related degradation features from degraded images. 

To address this challenge, we formulate multi-weather infrared image restoration as a conditional restoration process jointly guided by weather-type semantics and regional degradation severity. We propose a \textbf{Type-Severity Guided Progressive Disentanglement Network (TSGPD-IR)}. As shown in Fig.~1, TSGPD-IR contains three sequentially coupled modules. The \textbf{Weather and Semantic Co-Guided Multi-Level Prompt Generation Module (WS-MPG)} combines global weather-type semantics with stage-wise local features to generate adaptive prompts at different decoding stages. These prompts provide stage-specific guidance for suppressing weather artifacts and recovering thermal structures. The \textbf{Proxy-Supervised Regional Degradation Estimator (PS-RDE)} constructs proxy labels from paired degraded-clean regions, thereby learning the spatial distribution of degradation severity without manual severity annotations. Finally, the \textbf{Multi-Source Collaborative Expert Selection Strategy (MCESS)} performs expert routing by jointly leveraging weather-type semantics, regional degradation severity, and prompt-modulated stage features. Through this hierarchical collaborative process from weather-type to degradation severity, TSGPD-IR adaptively restores images affected by different weather conditions and local degradation levels within a unified framework.

\subsection{Weather and Semantic Co-Guided Multi-Level Prompt Generation Module}
Weather-induced responses in IR images are coupled with intrinsic thermal structures, making degradation guidance susceptible to scene-dependent thermal content. Existing prompt-based restoration methods \cite{2023promptir, 2026enhancing} typically derive degradation prompts from restoration features or employ degradation-conditioned prompts for feature modulation. However, feature-derived prompts may confuse weak thermal responses with weather interference, while stage-shared guidance cannot accommodate the distinct restoration requirements at different representation levels. To address this issue, as shown in Fig.~2, WS-MPG combines global weather semantics with stage-wise decoder features to generate adaptive prompts, progressively constraining the restoration from coarse degradation suppression to fine thermal-structure recovery.

\begin{figure}[pos=t]
\centering
\includegraphics[width=0.9\linewidth]{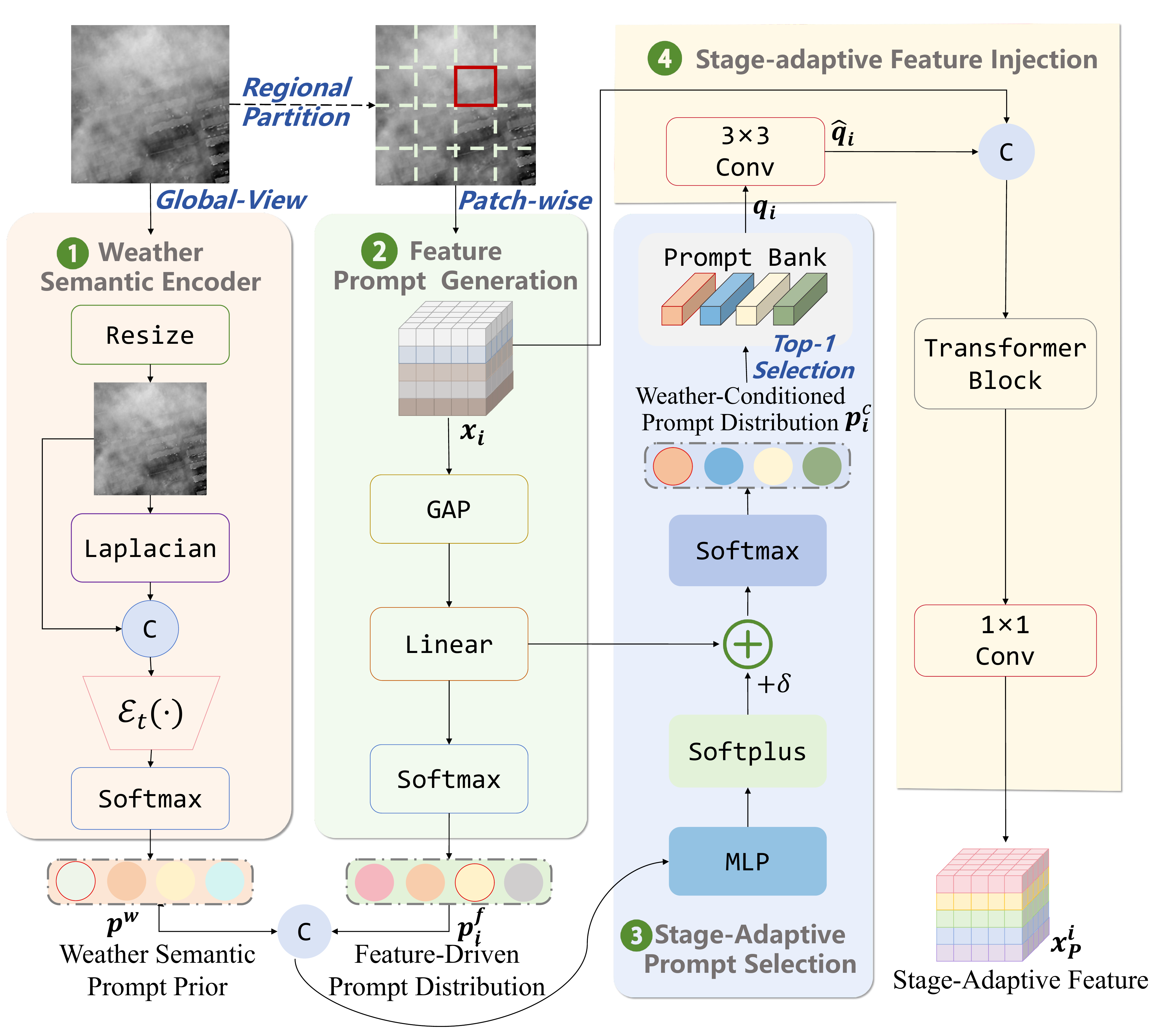}
\caption{Architecture of WS-MPG. The weather-semantic prior extracted from the degraded image and its Laplacian response calibrates the feature-driven prompt distribution at each decoding stage. The selected prompt is then injected into the corresponding decoder feature to provide progressive weather-aware guidance.}
\label{fig_2}
\end{figure}

Given a degraded infrared image $I_{d}$, we resize it to obtain a global task view $I_{task}$. To expose high-frequency cues related to edge attenuation and structural corruption, we concatenate $I_{task}$ with its Laplacian response to form the enhanced weather-aware input $I_{h}$:
\begin{equation}
I_{h}=Concat(I_{task},k_{Lap}*I_{task} ),
\end{equation}
where $k_{Lap}$ denotes the Laplacian convolution kernel. A weather semantic encoder $\mathcal{E}_t(\cdot)$ is employed to extract weather-type logits:
\begin{equation}
z=\mathcal{E}_t(I_{h}).
\end{equation}
A temperature-scaled Softmax is applied to obtain the weather semantic prompt prior:
\begin{equation}
p^w=Softmax\left(\frac{z}{\tau_{t}}\right),
\end{equation}
where $\tau_t$ is the temperature parameter. In addition to supporting weather-type classification supervision, $p^w$ serves as a global semantic prior for prompt modulation and subsequent expert routing.

At the $i$-th decoding stage, let $x_i\in\mathbb{R}^{C_i\times H_i\times W_i}$ denote the input feature. We first apply global average pooling to obtain a stage-level representation $f_i=GAP(x_i)$, which is mapped into the prompt-selection space: 
\begin{equation}
p^f_{i}=Softmax(W_{i}f_{i}+b_{i}),
\end{equation}
where $W_{i}$ and $b_{i}$ are learnable projection parameters. To explicitly incorporate weather semantics, we concatenate $p^f_{i}$ and $p^w$, and feed them into an MLP to generate positive prompt-wise scaling factors:
\begin{equation}
s_{i}=Softplus\left(MLP\left(Concat(p^f_{i},p^w)\right)\right)+\delta,
\end{equation}
where $Softplus(x)=\log(1+exp{(x)})$ and $\delta>0$ ensures positive scaling factors. Subsequently, to inject global weather semantics into the local prompt-selection space, $s_{i}$ is used to calibrate the local prompt-routing logits, yielding a weather-conditioned prompt distribution:
\begin{equation}
p_{i}^{c}=Softmax(W_{i}f_{i}+b_{i}+\log s_{i}).
\end{equation}

To improve prompt specificity, we adopt top-1 sparse prompt selection. Let the prompt bank at the $i$-th stage be defined as $P_{i}=\{P_{i}^{n}\}_{n=1}^{N}$, where $P_{i}^{n}\in\mathbb{R}^{C_i \times H_i \times W_i}$ and $N$ denotes the number of candidate prompts. The selected prompt is computed as: 
\begin{equation}
q_{i}=\sum_{n=1}^{N}\omega_{i,n}P_{i}^{n},
\ \ \omega_{i}=SparseTopK(p_{i}^{c},1),
\end{equation}
where $\omega_{i,n}$ denotes the sparse activation weight and is defined as:
\begin{equation}
\omega_{i,n}=\begin{cases}1, & n=\arg\max\limits_m p_{i}^{c,m}, \\
0, & \text{otherwise}.\end{cases}
\end{equation}
Subsequently, through scale alignment and local convolutional mapping, $q_{i}$ is transformed into a stage-adaptive prompt feature: $\widehat{q}_{i}=Conv_{3\times3}\left(q_{i}\right)$. It is then injected into the corresponding decoder stage through feature concatenation and transformer-based fusion: 
\begin{equation}
x_{P}^{i}=Conv_{1\times1}\left(TransformerBlock\left(Concat(x_{i},\widehat{q}_{i})\right)\right).
\end{equation}

A single-stage prompt is insufficient to reliably distinguish weather-induced responses from intrinsic thermal structures across different representation levels. Therefore, WS-MPG progressively injects weather-semantic guidance from deep to shallow decoding stages, enabling coarse-to-fine separation of degradation interference from real thermal content. Specifically, deep-stage prompts regulate the global restoration direction, intermediate-stage prompts facilitate thermal-contrast recovery and structural reconstruction, and shallow-stage prompts refine edges and fine-grained thermal responses.
\begin{figure*}[h]
\centering
\includegraphics[width=0.9\textwidth]{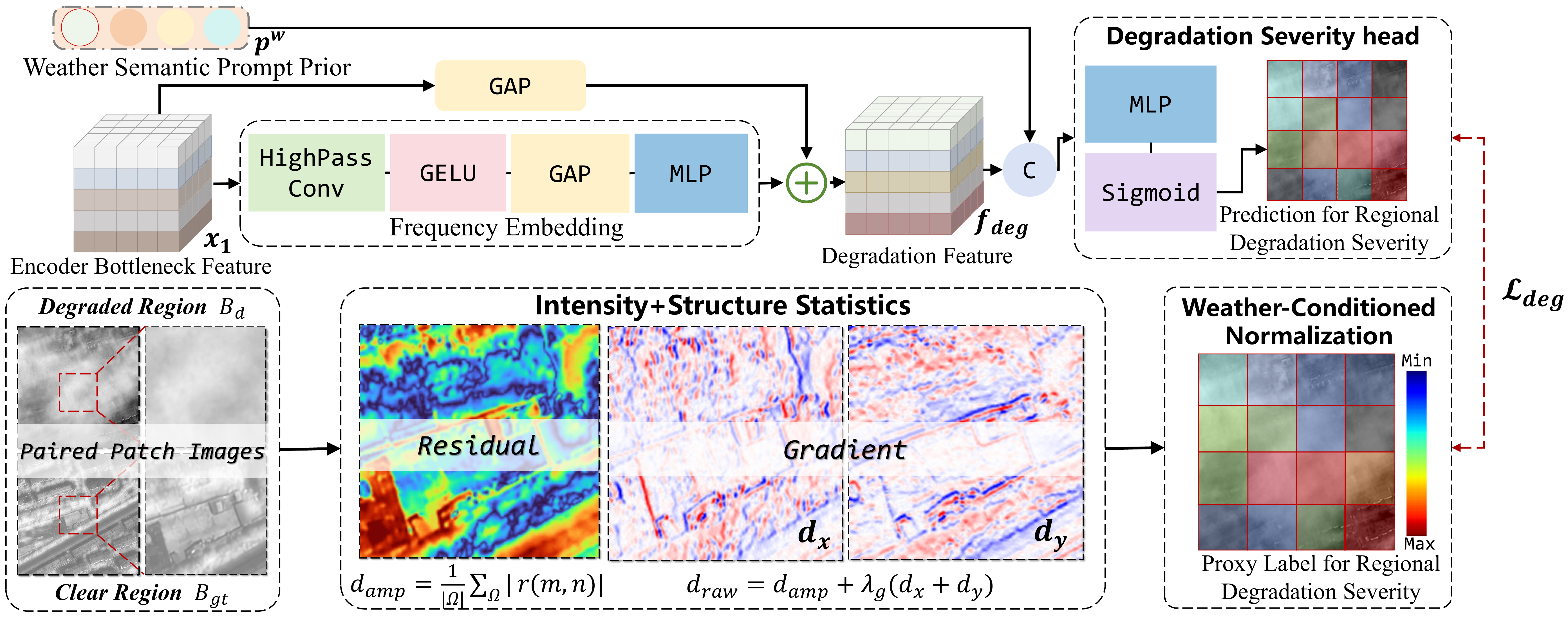}
\caption{Architecture of PS-RDE. For each aligned degraded and clear region pair, residual magnitude and gradient statistics are combined and weather-conditionally normalized to construct a regional severity proxy. The prediction branch estimates the corresponding severity from the degraded-region feature, frequency embedding, and weather-semantic prior.}
\label{fig_3}
\end{figure*}

\subsection{Proxy-Supervised Regional Degradation Estimator}
Adverse-weather degradation in IR images is often spatially nonuniform, with different regions exhibiting different degrees of occlusion and thermal-contrast attenuation. Existing strategies mainly rely on global degradation information or degradation-entangled features, making it difficult to explicitly characterize regional severity variations within the same image. To address this problem, we develop PS-RDE to explicitly estimate regional degradation severity. As illustrated in Fig.~\ref{fig_3}, PS-RDE constructs proxy severity supervision from paired degraded and clean images and further performs weather-conditioned normalization. The resulting severity prior guides subsequent expert routing to adapt the restoration capacity to local degradation conditions.

Given a spatially aligned clear region $B_{gt}$ and degraded region $B_{d}$, the degradation residual magnitude $|r|$ reflects the overall shift and intensity distributions caused by weather degradation. Let $\Omega$ denote the spatial domain of the current region, and let $|\Omega|$ denote the number of valid pixels. The mean residual magnitude $d_{amp}$ is defined as:
\begin{equation}
r=B_{gt}-B_{d},\ \ d_{amp}=\frac{1}{|\Omega|}\sum_{(m,n)\in\Omega}|r(m,n)|.
\end{equation}

In contrast, gradient variations reflect structural degradation, and we define horizontal and vertical gradient differences as:
\begin{equation}
d_{x} = \frac{1}{|\Omega|}\sum_{(m,n)\in\Omega}|r(m+1,n)-r(m,n)|,
\end{equation}

\begin{equation}
d_{y}= \frac{1}{|\Omega|}\sum_{(m,n)\in\Omega}|r(m,n+1)-r(m,n)|.
\end{equation}
We combine intensity deviation and structural perturbation to define the raw degradation severity:
\begin{equation}
d_{raw}=d_{amp}+\lambda_g\left(d_x+d_y\right),
\end{equation}
where $\lambda_g$ balances the contributions of intensity discrepancy and local structural distortion.

However, the same residual magnitude may correspond to different degradation levels under different weather conditions. For example, haze typically produces smooth, large-scale thermal-contrast compression, whereas cloud occlusion often causes localized obscuration and structural loss. We therefore perform weather-conditioned normalization, and let $\mu_c$ and $\sigma_c$ denote the mean and standard deviation of $d_{\mathrm{raw}}$ over training samples belonging to weather category $c$. The normalized proxy label for regional degradation severity is then defined as:
\begin{equation}
d_{proxy}=Sigmoid\left(\frac{d_{\mathrm{raw}}-\mu_c}{\tau_d\sigma_c+\epsilon}\right),
\end{equation}
where $\tau_d$ denotes the severity-normalization temperature coefficient, and $\epsilon$ is a numerical stability constant. This normalization suppresses type-dependent response shifts and enables the proxy label to emphasize relative degradation severity within each weather.

To estimate regional severity from the degraded input, we construct a weather-aware regional degradation severity prediction branch. Let $x_{1}$ denote the encoder bottleneck feature. We first obtain a global semantic representation through global average pooling: $e_{g}=GAP(x_{1})$. To complement semantic features with degradation-sensitive frequency cues, we introduce a frequency embedding module: $e_{f}=\mathcal{E}_{freq}(x_{1})$. The resulting degradation-aware representation is given by: $f_{deg}=e_{g}+e_{f}$. In addition, $f_{deg}$ is further concatenated with $p^w$ and fed into a prediction head to estimate the severity of regional degradation:
\begin{equation}
d_{pred}=Sigmoid\left(MLP\left(Concat(f_{deg},p^w)\right)\right).
\end{equation}

By leveraging paired data to derive proxy supervision and explicitly conditioning severity estimation on weather semantics, PS-RDE disentangles weather type from regional degradation intensity. The predicted severity is subsequently used for expert routing, enabling finer-grained and more adaptive infrared image restoration.

\subsection{Multi-Source Collaborative Expert Selection Strategy}
\begin{figure}[pos=t]
\centering
\includegraphics[width=\linewidth]{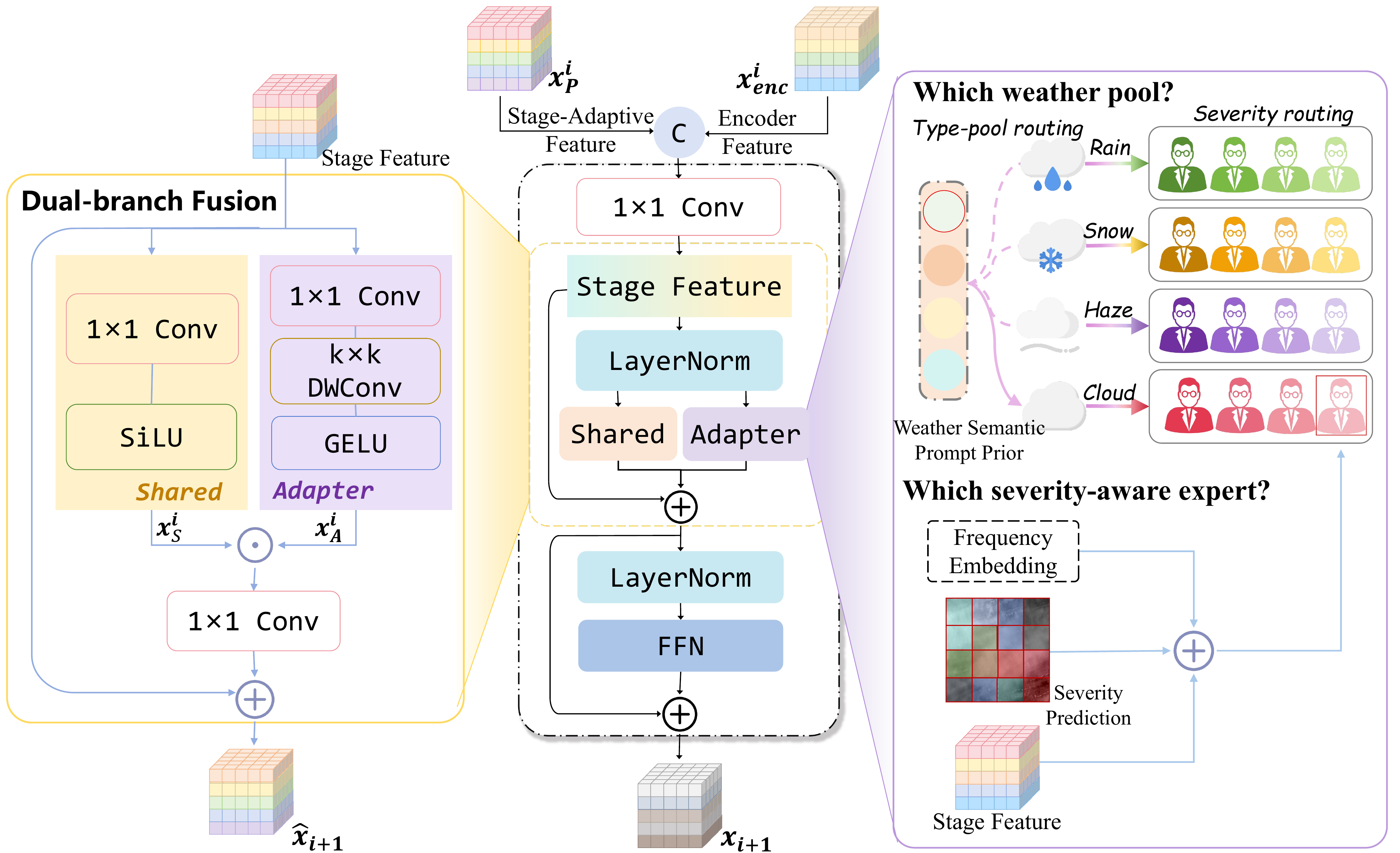}
\caption{Architecture of MCESS. The shared branch extracts weather-invariant restoration features, while the adaptive branch first selects a weather-specific expert pool and then routes each region to a severity-compatible expert. Their outputs are fused with the prompt-modulated stage feature for adaptive restoration.}
\label{fig_4}
\end{figure}

Weather type determines the dominant restoration direction, whereas regional severity determines the required local restoration capacity. Existing degradation-aware expert models \cite{mofme,2025complexity} commonly perform routing over a shared expert space, where different degradation types may compete for the same experts and spatially varying regions may receive insufficiently differentiated restoration. To translate type and severity priors into restoration behaviors, MCESS decomposes routing into two hierarchical decisions. As shown in Fig.~\ref{fig_4}, weather semantics first activate a task-specific expert pool, and regional severity subsequently selects a severity-compatible expert. A shared branch further preserves weather-invariant thermal structures. In this way, MCESS can select the restoration direction according to weather type and adjust the restoration strength according to local degradation conditions.

The shared branch is designed to recover weather-invariant thermal structures. Let $x_P^{i}$ denote the prompt-modulated feature at the $i$-th decoding stage. After layer normalization $LN(\cdot)$, a shared attention branch $\mathcal{A}_{s}(\cdot)$ extracts generic restoration features: $x_{S}^{i}=\mathcal{A}_{s}\left(LN(x_{P}^{i})\right)$, where $\mathcal{A}_{s}(\cdot)$ consists of channel projection and shared attention transformation. By learning common structural priors across weather conditions, this branch provides stable basic restoration capability and reduces redundancy among weather-specific experts.

To separate global weather-type selection from local restoration-strength adjustment, building on the shared branch, we introduce a hierarchical type-severity routing mechanism. Specifically, weather semantics determine task-level restoration directions, whereas regional degradation severity controls the restoration capacity within each task pool. 

For task-level routing, we use top-1 sparse selection based on the weather semantic prior $p^w$. Given weather-specific task pools, each representing a weather-dominated restoration subspace, the task-level routing weights are computed as:
\begin{equation}
w_{task}=SparseTopK(p^w,1),
\end{equation}
Thus, weather-type recognition is directly coupled with task-pool selection, preventing unconstrained competition among different weather degradations in the expert space.

However, weather type alone cannot characterize the spatial non-uniformity of degradation within the same image. Therefore, each task pool further contains multiple complexity-level experts with progressively increasing restoration capacity. Let $c_l$ denote the severity center of the $l$-th complexity expert, and the severity-aware bias for the $l$-th expert is formulated as $b_{diff}^{l}(d_{pred})=-\beta\left(d_{pred}-c_l\right)^2$, where $\beta$ controls the strength of the severity constraint. This bias increases the routing response of the expert whose severity center is closest to $d_{\mathrm{pred}}$. The routing logit of the $l$-th expert at the $i$-th decoding
stage is defined as:
\begin{equation}
g_{i,l} = w_{g, l}\cdot GAP\left(x_{P}^{i}\right)+w_{f,l}\cdot e_{f}+b_{diff}^{l}\left(d_{pred}\right),
\end{equation}
where $e_f$ is the bottleneck frequency embedding, and ${w}_{g,l}$ and ${w}_{f,l}$ are learnable projection parameters. The complexity-level routing distribution is obtained using temperature-scaled Softmax: $\widehat{w}_{level}^{(i)}=Softmax\left({{g}_i}/{\tau_r}\right)$, where $\tau_r$ denotes routing temperature parameter. To enforce explicit expert specialization during restoration, we further apply top-1 sparse selection:
\begin{equation}
{w}_{level}^{(i)}=SparseTopK\left(\widehat{w}_{level}^{(i)},1\right),
\end{equation}
Accordingly, the task-level and severity-level routing weights form a two-dimensional gating mechanism. The joint gate for the $l$-th complexity expert in the $t$-th task pool is defined as:
\begin{equation}
{G}_{i,t,l}=w_{task}{(t)}\cdot{w}_{level}^{(i)}{(l)}.
\end{equation}
Let $\mathcal{R}_{t,l}(\cdot)$ denote the $l$-th complexity expert in the $t$-th weather-specific expert pool. The output of the expert restoration branch is given by:
\begin{equation}
x_A^i=\sum_{t=1}^{T}\sum_{l=1}^{L}G_{i,t,l}\mathcal{R}_{t,l}(x_P^i).
\end{equation}
This routing scheme selects a weather-compatible restoration direction at the task level and an appropriate restoration capacity at the severity level.

Finally, the prompt-modulated feature, shared restoration feature, and expert restoration feature are fused as: $\widehat{x}_{i+1}=x_{P}^{i}+x_{S}^{i}+x_{A}^{i}$. The fused feature is further refined by a feed-forward enhancement module $\mathcal{F}_{\mathrm{ffn}}$ to produce the output of the current decoding block:
\begin{equation}
x_{i+1}=\widehat{x}_{i+1}+\mathcal{F}_{\mathrm{ffn}}\left(LN(\widehat{x}_{i+1})\right).
\end{equation}

As the level $l$ increases, the experts are progressively strengthened in terms of convolutional kernel size, low-rank channel capacity, and residual-stack depth, thereby forming a hierarchy of restoration capacities from low to high. Lightweight experts are better suited to mildly degraded regions, whereas high-complexity experts are more appropriate for severely degraded regions. This expert selection mechanism effectively alleviates negative transfer across different weather degradations and improves the interpretability of expert specialization.

\subsection{Training Strategy and Optimization Objectives}
To prevent unreliable weather-semantic and degradation-severity estimates from destabilizing expert routing in early training, we adopt a two-stage optimization strategy. Stage I learns stable weather-semantic prompts and aligns them with prompt-guided restoration behaviors. Stage II introduces proxy-supervised degradation estimation and hierarchical type-severity collaborative expert routing. This progressive design establishes conditional restoration from global weather semantics to local restoration capacity.

\subsubsection{Stage I: Weather-Semantic Prior Learning and Prompt Alignment}
In Stage I, we jointly train the weather-type encoder, WS-MPG, and the coupled restoration backbone to learn stable weather-semantic representations. The objective combines pixel reconstruction, weather classification, and contrastive prompt restoration losses:
\begin{equation}
\mathcal{L}_{s1}=\mathcal{L}_{\mathrm{pix}}+\lambda_{\mathrm{ce}}\mathcal{L}_{\mathrm{ce}}+\lambda_{\mathrm{cpr}}\mathcal{L}_{\mathrm{CPR}}.
\end{equation}

The pixel reconstruction loss enforces pixel-wise consistency between the restored image $I_e$ and its ground truth $I_{gt}$: $\mathcal{L}_{\mathrm{pix}}=\|I_{e}-I_{gt}\|_{1}$. The weather classification loss supervises the weather-type encoder to produce discriminative degradation-type representations: $\mathcal{L}_{\mathrm{ce}}=CE(z,y)$, where $z$ denotes the predicted weather logits and $y$ is the weather-type label.

To ensure that weather prompts actively influence restoration, we introduce a contrastive prompt restoration loss. Given the restoration result $I_{pos}$ produced under the correct prompt and the results ${I_{\mathrm{neg}}^{(j)}}$ generated under incorrect prompts, the perceptual distance to $I_{gt}$ is defined as: $D(I_{pos},I_{gt})=\sum_{m}\left\|\phi_{m}(I_{pos})-\phi_{m}(I_{gt})\right\|_{2}^{2}$,
where $\phi_{m}(\cdot)$ denotes the feature response of the fixed perceptual feature extractor at the $m$-th layer. The contrastive prompt restoration loss is then formulated as:
\begin{equation}
\mathcal{L}_{\mathrm{CPR}}=D(I_{pos},I_{gt})-\frac{\lambda_{\mathrm{neg}}}{N} \sum_{j=1}^{N}D(I_{\mathrm{neg}}^{(j)},I_{gt}),
\end{equation}
where $N$ denotes the number of negative prompt samples. Minimizing this loss encourages the correct prompt to yield a restoration result closer to the ground truth, while suppressing interference caused by mismatched prompts. 

Stage-I training provides a stable semantic initialization for severity-aware routing in Stage II.

\subsubsection{Stage II: Degradation-Aware Dynamic Routing Optimization}
Based on the weather-semantic priors learned in Stage I, Stage II incorporates PS-RDE and MCESS to enable degradation-severity-aware dynamic restoration. The overall objective is:
\begin{equation}
\mathcal{L}_{s2}=\mathcal{L}_{\mathrm{pix}}+\mathcal{L}_{\mathrm{CPR}}+\lambda_{\mathrm{deg}}\mathcal{L}_{\mathrm{deg}}+\lambda_{\mathrm{cons}}\mathcal{L}_{\mathrm{cons}}+\lambda_{\mathrm{moe}}\mathcal{L}_{\mathrm{moe}}.
\end{equation}

According to Section~3.3, the proxy supervision loss for degradation estimation is formulated as: 
\begin{equation}
\mathcal{L}_{\mathrm{deg}}=SmoothL1(d_{pred},d_{proxy}).
\end{equation}.

To ensure consistency between regional degradation severity prediction and implicit expert selection, we define the severity expectation induced by the complexity-level routing weights at the $i$-th decoding stage as: $d_{route}^{(i)}=
\sum_{l=1}^{L}\hat{w}_{level}^{i}{(l)}c_{l}$. The consistency loss is defined as:
\begin{equation}
\mathcal{L}_{\mathrm{cons}}=\frac{1}{S}\sum_{i=1}^{S}SmoothL1(d_{route}^{(i)},d_{pred}),
\end{equation}
where $S$ denotes the number of decoding stages. This constraint ensures that degradation prediction is supervised not only by proxy labels but also by the restoration strength implied by complexity-level routing.

To alleviate the persistent overuse of a small subset of complexity experts, we further introduce an auxiliary MoE balancing loss. Since the task-level routing weight $w_{task}$ is determined by $p^w$, the balancing operation is performed over complexity-level experts within each activated task pool. For a mini-batch $\mathcal{X}$, the set of samples activating the $t$-th task pool is $\mathcal{X}_t=\{ {\mathbf{x}\in\mathcal{X}\mid w_{task}{(t;\mathbf{x})}>0 \}}$.

For the $l$-th complexity expert $\mathcal{R}_{t,l}$ in task pool $\mathcal{R}_{t}$, let $n_l$ denote its number of learnable parameters. We introduce a complexity bias: $b_{l}={n_l}/{n_{\max}}$, where $n_{max}$ is the maximum number of learnable parameters among all complexity experts. At the $i$-th decoding stage, the complexity-weighted importance of the $l$-th expert in the $t$-th task pool is defined as:
\begin{equation}
\mathrm {Imp}_{t,l}^{(i)}=\frac{b_l}{|\mathcal{X}_t|}\sum_{x\in\mathcal{X}_t}
\widehat{w}_{level}^{(i)}\left(l;x\right).
\end{equation}
The corresponding complexity-weighted load is defined under top-1 hard routing as:
\begin{equation}
\mathrm {Load}_{t,l}^{(i)}=\frac{b_l}{|\mathcal{X}_t|}\sum_{x\in\mathcal{X}_t}\mathbb{I}[l=\arg\max_{j}\widehat{w}_{level}^{(i)}\left(j;x\right) ],
\end{equation}
where $\mathbb{I}[\cdot]$ is the indicator function. Let $\mathrm{Imp}_{t}^{(i)}=[{\mathrm{Imp}_{t,l}^{(i)}}]_{l=1}^{L}$ and $\mathrm{Load}_{t}^{(i)}=[{\mathrm{Load}_{t,l}^{(i)}}]_{l=1}^{L}$. The final auxiliary MoE balancing loss is defined as:
\begin{align}\label{equ}
\nonumber
\mathcal{L}_{\mathrm{moe}} = \frac{1}{S}\sum_{i=1}^{S}\frac{1}{N_t} \sum_{t:\mathcal{X}t\neq\emptyset}\Bigg[\frac{1}{2}&\mathrm{CV}\left(
\mathrm{Imp}_{t}^{(i)}\right)^2\\ 
+\frac{1}{2}&\mathrm{CV}\left(\mathrm{Load}_{t}^{(i)}\right)^2\bigg]
\end{align}
where $N_t$ denotes the number of non-empty task pools, and $\mathrm{CV}(\cdot)$ denotes the coefficient of variation. This loss encourages a more balanced use of complexity experts while preserving their functional specialization, thereby alleviating expert collapse.

With two-stage training, the model first learns a stable weather-semantic prior and then learns region-wise severity estimation and type–severity routing.

\section{Experiments}
\subsection{Implementation Details}
\subsubsection{Training Settings}
During training, we randomly crop $160 \times 160$ patches as inputs to the restoration backbone. Meanwhile, each degraded input is resized to $192 \times 192$ to construct a task-level view, which is then fed into the weather semantic encoder. Random horizontal flipping and rotation are employed for data augmentation.

The network contains four weather-specific task pools, each comprising four experts with different complexity levels. We adopt the AdamW optimizer with an initial learning rate of $2 \times 10^{-4}$. A cosine-annealing learning-rate schedule is employed, with the minimum learning rate set to $1 \times 10^{-7}$. The batch size is set to 8. The number of negative-prompt samples is set to 3, while $\lambda_{neg}$ is set to 0.01. The weights of the degradation-severity supervision loss and the severity-consistency loss are set to 0.02 and 0.01, respectively. The weight of the MoE load-balancing loss $\lambda_{moe}$ is set to 1.0. All experiments are conducted on four NVIDIA RTX 2080 GPUs. 
\begin{table*}[pos=t]
\centering
\caption{Comprehensive evaluation on the WeatherIR benchmark. The best results are highlighted in \textbf{bold}, while the second-best results are \underline{underlined}.}
\label{tab:quantitative_comparison}
\setlength{\tabcolsep}{2pt}
\renewcommand{\arraystretch}{1.4}
\resizebox{\textwidth}{!}{\begin{tabular}{c|l|ccc|ccc|ccc|ccc|ccc}
\toprule
\multirow{2}{*}{\textbf{Type}} & \multirow{2}{*}{\textbf{Method (Venue)}} & \multicolumn{3}{c|}{\textbf{Rain}} & \multicolumn{3}{c|}{\textbf{Haze}} & \multicolumn{3}{c|}{\textbf{Snow}} & \multicolumn{3}{c|}{\textbf{Cloud}} & \multicolumn{3}{c}{\textbf{Average}} \\
\cline{3-17}
& & PSNR$\uparrow$ & SSIM$\uparrow$ & LPIPS$\downarrow$ & PSNR$\uparrow$ & SSIM$\uparrow$ & LPIPS$\downarrow$ & PSNR$\uparrow$ & SSIM$\uparrow$ & LPIPS$\downarrow$ & PSNR$\uparrow$ & SSIM$\uparrow$ & LPIPS$\downarrow$ & PSNR$\uparrow$ & SSIM$\uparrow$ & LPIPS$\downarrow$ \\
\midrule
\multirow{5}{*}{\rotatebox[origin=c]{90}{\textit{General}}}
& Restormer \cite{2022restormer} (CVPR'22) & 25.7891 & 0.8919 & 0.1430 & 26.4475 & 0.9545 & \underline{0.0571} & 20.8223 & 0.7150 & 0.2989 & \textbf{25.9496} & \textbf{0.7975} & \underline{0.2595} & 24.7521 & 0.8397 & 0.1896 \\
& CODE \cite{2023code} (CVPR'23) & 25.1391 & 0.8875 & 0.1328 & \textbf{27.8787} & \underline{0.9597} & 0.0614 & 17.2871 & 0.6354 & 0.3215 & \underline{25.8981} & 0.7893 & 0.3015 & 24.0508 & 0.8180 & 0.2043 \\
& MambaIR \cite{2024mambair} (ECCV'24) & 25.7942 & 0.8913 & 0.1437 & 26.3540 & 0.9548 & 0.0806 & 21.6954 & 0.7927 & 0.2027 & 25.1563 & 0.7712 & 0.3019 & 24.7500 & 0.8525 & 0.1822 \\
& VmambaIR \cite{2025vmambair} (TCSVT'25) & 25.7905 & 0.8934 & 0.1267 & 25.5185 & 0.9403 & 0.0867 & 24.9894 & 0.8557 & 0.1731 & 25.2243 & 0.7831 & 0.2606 & 25.3807 & 0.8681 & 0.1618 \\
& MaIR \cite{2025mair} (CVPR'25) & 25.8046 & 0.8918 & 0.1501 & 25.3824 & 0.9320 & 0.1116 & 25.1223 & 0.8757 & 0.1785 & 25.6610 & 0.7919 & 0.2806 & 25.4926 & \underline{0.8729} & 0.1802 \\
\midrule
\multirow{6}{*}{\rotatebox[origin=c]{90}{\textit{All-in-One}}}
& PromptIR \cite{2023promptir} (NeurIPS'23) & \underline{30.3100} & \underline{0.9496} & \underline{0.0620} & 23.8900 & 0.9373 & 0.0848 & \underline{28.7800} & \underline{0.9197} & \underline{0.0819} & 23.4800 & 0.6699 & 0.3827 & \underline{26.6150} & 0.8691 & \underline{0.1529} \\
& InstructIR \cite{2024instructir} (ECCV'24) & 26.5600 & 0.8994 & 0.1414 & 24.4281 & 0.9461 & 0.0725 & 25.1898 & 0.8695 & 0.1620 & 23.1149 & 0.7112 & 0.4460 & 24.8232 & 0.8566 & 0.2055 \\
& PPFN \cite{2026enhancing} (NeurIPS'25) & 19.9626 & 0.2086 & 0.4395 & 13.6959 & 0.4106 & 0.3598 & 19.1910 & 0.1678 & 0.4585 & 13.6479 & 0.4756 & 0.3690 & 16.6244 & 0.3157 & 0.4067 \\
& CPL \cite{2025beyond} (TPAMI'25) & 26.9040 & 0.8987 & 0.0932 & 22.8535 & 0.9315 & 0.0941 & 26.1484 & 0.8760 & 0.1151 & 22.7093 & 0.6966 & 0.3490 & 24.6538 & 0.8507 & 0.1629 \\
& MoCE-IR \cite{2025complexity} (CVPR'25) & 28.8431 & 0.9354 & 0.0797 & 23.0352 & 0.9266 & 0.1251 & 27.3740 & 0.9147 & 0.0974 & 23.3285 & 0.7045 & 0.4044 & 25.6452 & 0.8703 & 0.1767 \\
\cline{2-17}
& \cellcolor{red!15} \textbf{Ours} & \cellcolor{red!15} \textbf{33.4397} & \cellcolor{red!15} \textbf{0.9714} & \cellcolor{red!15} \textbf{0.0311} & \cellcolor{red!15} \underline{26.6045} & \cellcolor{red!15} \textbf{0.9625} & \cellcolor{red!15} \textbf{0.0477} & \cellcolor{red!15} \textbf{32.5130} & \cellcolor{red!15} \textbf{0.9582} & \cellcolor{red!15} \textbf{0.0409} & \cellcolor{red!15} 25.7383 & \cellcolor{red!15} \underline{0.7929} & \cellcolor{red!15} \textbf{0.2490} & \cellcolor{red!15} \textbf{29.5739} & \cellcolor{red!15} \textbf{0.9213} & \cellcolor{red!15} \textbf{0.0922} \\
\bottomrule
\end{tabular}}
\end{table*}
\begin{table*}[pos=t]
\centering
\caption{Task-specific quantitative comparisons on four single-degradation infrared image restoration tasks. The best and second-best results are highlighted in \textbf{bold} and \underline{underlined}, respectively.}
\label{tab:task_specific_comparison}
\footnotesize
\setlength{\tabcolsep}{3.5pt}
\renewcommand{\arraystretch}{0.95}
\begin{minipage}[pos=t]{0.49\textwidth}
\centering
\textit{(a) Deraining}\par
\vspace{1pt}
\begin{tabularx}{\linewidth}{@{}>{\raggedright\arraybackslash}>{\centering\arraybackslash}p{0.42\linewidth}*{3}{>{\centering\arraybackslash}X}@{}}
\toprule
Method (Venue) & PSNR$\uparrow$ & SSIM$\uparrow$ & LPIPS$\downarrow$ \\
\midrule
IDT~\cite{2022idt} (TPAMI'22) & \underline{25.8044} & 0.8877 & 0.1441 \\
DRSformer~\cite{2023drsformer} (CVPR'23)  & 25.7816 & \underline{0.8913} & \textbf{0.1397} \\
HCT-FFN~\cite{2023hctffn} (AAAI'23)  & 25.7848 & 0.8904 & 0.1540 \\
NeRD-Rain~\cite{2024nerdrain} (CVPR'24)  & 25.0497 & 0.8698 & 0.1664 \\
CSUD~\cite{2025csud} (CVPR'25)  & 24.3442 & 0.8006 & 0.1637 \\
\midrule
\rowcolor{red!10}
\textbf{Ours} & \textbf{25.8240} & \textbf{0.8928} & \underline{0.1426} \\
\bottomrule
\end{tabularx}
\end{minipage}
\hfill
\begin{minipage}[pos=t]{0.49\textwidth}
\centering
\textit{(b) Dehazing}\par
\vspace{1pt}
\begin{tabularx}{\linewidth}{@{}>{\raggedright\arraybackslash}>{\centering\arraybackslash}p{0.42\linewidth}*{3}{>{\centering\arraybackslash}X}@{}}
\toprule
Method (Venue) & PSNR$\uparrow$ & SSIM$\uparrow$ & LPIPS$\downarrow$ \\
\midrule
AOD-Net~\cite{2017aod} (ICCV'17)  & 12.3522 & 0.6699 & 0.4884 \\
DehazeFormer~\cite{2023dehazeformer} (TIP'23)   & 25.9021 & \underline{0.9510} & \underline{0.0641} \\
PBD~\cite{2025pbd} (TCSVT'25) & 13.0654 & 0.6824 & 0.2622 \\
DehazeXL~\cite{2025dehazexl} (CVPR'25)  & 22.6899 & 0.9078 & 0.1202 \\
BiLaLoRA~\cite{2026bilalora} (CVPR'26)  & \underline{27.5843} & 0.9432 & 0.0820 \\
\midrule
\rowcolor{red!10}
\textbf{Ours} & \textbf{29.6695} & \textbf{0.9689} & \textbf{0.0371} \\
\bottomrule
\end{tabularx}
\end{minipage}
\par\vspace{0.8em}
\begin{minipage}[pos=t]{0.49\textwidth}
\centering
\textit{(c) Desnowing}\par
\vspace{1pt}
\begin{tabularx}{\linewidth}{@{}>{\raggedright\arraybackslash}>{\centering\arraybackslash}p{0.42\linewidth}*{3}{>{\centering\arraybackslash}X}@{}}
\toprule
Method (Venue) & PSNR$\uparrow$ & SSIM$\uparrow$ & LPIPS$\downarrow$ \\
\midrule
DesnowNet~\cite{2018desnownet} (TIP'18)   & 16.6453 & 0.6591 & 0.4312 \\
DDMSNet~\cite{2021ddmsnet} (TIP'21)   & 17.5991 & 0.5405 & 0.5481 \\
HDCWNet~\cite{2021hdcwnet} (ICCV'21)  & 21.1716 & 0.7598 & 0.2692 \\
InvDSNet~\cite{2023invdsnet} (TCSVT'23) & 22.1677 & 0.8082 & 0.2252 \\
SMGARN~\cite{2023smgarn} (CVIU'23)  & \underline{25.0570} & \underline{0.8683} & \underline{0.1906} \\
\midrule
\rowcolor{red!10}
\textbf{Ours} & \textbf{25.4031} & \textbf{0.8862} & \textbf{0.1463} \\
\bottomrule
\end{tabularx}
\end{minipage}
\hfill
\begin{minipage}[pos=t]{0.49\textwidth}
\centering
\textit{(d) Cloud Removal}\par
\vspace{1pt}
\begin{tabularx}{\linewidth}{@{}>{\raggedright\arraybackslash}>{\centering\arraybackslash}p{0.4\linewidth}*{3}{>{\centering\arraybackslash}X}@{}}
\toprule
Method (Venue) & PSNR$\uparrow$ & SSIM$\uparrow$ & LPIPS$\downarrow$ \\
\midrule
MRF-Net~\cite{2024mrfnet} (TGRS'24) & 20.0971 & 0.6223 & 0.2912 \\
CMNet~\cite{2024cmnet} (TGRS'24) & 25.6479 & 0.7857 & 0.2958 \\
CR-Former~\cite{2024crformer} (TGRS'24) & \underline{25.8260} & \underline{0.7933} & 0.2704 \\
EMRDM~\cite{2025emrdm} (CVPR'25) & 23.3522 & 0.7391 & \underline{0.2332} \\
CR-Famba~\cite{2025crfamba} (TMM'25)  & 24.3412 & 0.7313 & 0.4040 \\
\midrule
\rowcolor{red!10}
\textbf{Ours} & \textbf{26.9701} & \textbf{0.8181} & \textbf{0.2177} \\
\bottomrule
\end{tabularx}
\end{minipage}
\end{table*}

\subsubsection{Evaluation Metrics}
We employ the peak signal-to-noise ratio (PSNR), structural similarity index measure (SSIM) \cite{2004ssim}, and Learned Perceptual Image Patch Similarity (LPIPS) \cite{2018lpips} to assess the quality of the restored images. PSNR and SSIM primarily measure pixel-level fidelity and structural similarity, respectively, with higher values indicating better restoration performance. LPIPS evaluates perceptual discrepancy between the restored and reference images, for which a lower value indicates better perceptual quality. 
\subsubsection{Datasets}
We evaluate TSGPD-IR on public datasets covering four categories of weather degradation. Specifically, we employ AWMM-100K \cite{2026awmm}, the M3FD-Synthetic dataset (M3FD\_TIR) \cite{2026tsfanet}, and two subsets of the CUHK-CR dataset \cite{2024cuhk}, namely CUHK-CR1 and CUHK-CR2. AWMM-100K provides samples degraded by rain, haze, and snow, whereas M3FD\_TIR contains haze-degraded thermal infrared images. CUHK-CR1 and CUHK-CR2 contain samples affected by thin and thick clouds, respectively. For unified multi-weather restoration, we combine samples from the datasets above to build a mixed dataset called WeatherIR, which covers four types of degradation. All collected samples are randomly divided into training and test sets at a ratio of 9:1.
\subsection{Comparison With State-of-the-Art Methods}
In this section, we conduct extensive experiments under two settings to validate the effectiveness of the proposed TSGPD-IR framework: (a) all-in-one and (b) Task-specific. In the all-in-one setting, a unified model is trained to handle multiple degradation types and is evaluated on four categories of weather degradation. In contrast, the task-specific setting involves training separate models, each specialized for a specific restoration task.
\subsubsection{Four-Task All-in-One Restoration}
\noindent\textbf{Quantitative Analysis:}
We compare the TSGPD-IR with five representative all-in-one methods and five general-purpose image restoration methods.  As shown in Table~\ref{tab:quantitative_comparison}, TSGPD-IR achieves the best average performance, with a PSNR of 29.5739~dB, an SSIM of 0.9213, and an LPIPS of 0.0922. Compared with PromptIR, which obtains the second-best average PSNR and LPIPS, TSGPD-IR improves PSNR by 2.9589~dB and reduces LPIPS by 39.70\%. It also improves the average SSIM by 5.54\% over MaIR. These results demonstrate the effectiveness of TSGPD-IR in jointly handling infrared weather degradations. For rain and snow removal, TSGPD-IR ranks first across all three metrics, indicating its strong capability to remove spatially distributed degradation while preserving thermal structures. For dehazing, it achieves the best SSIM and LPIPS together with the second-best PSNR, while for cloud removal it obtains the lowest LPIPS and the second-best SSIM. These results indicate that TSGPD-IR maintains a favorable balance between pixel fidelity, structural consistency, and perceptual quality across heterogeneous weather degradations.

\begin{figure*}[pos=t]
\centering
\includegraphics[width=\textwidth]{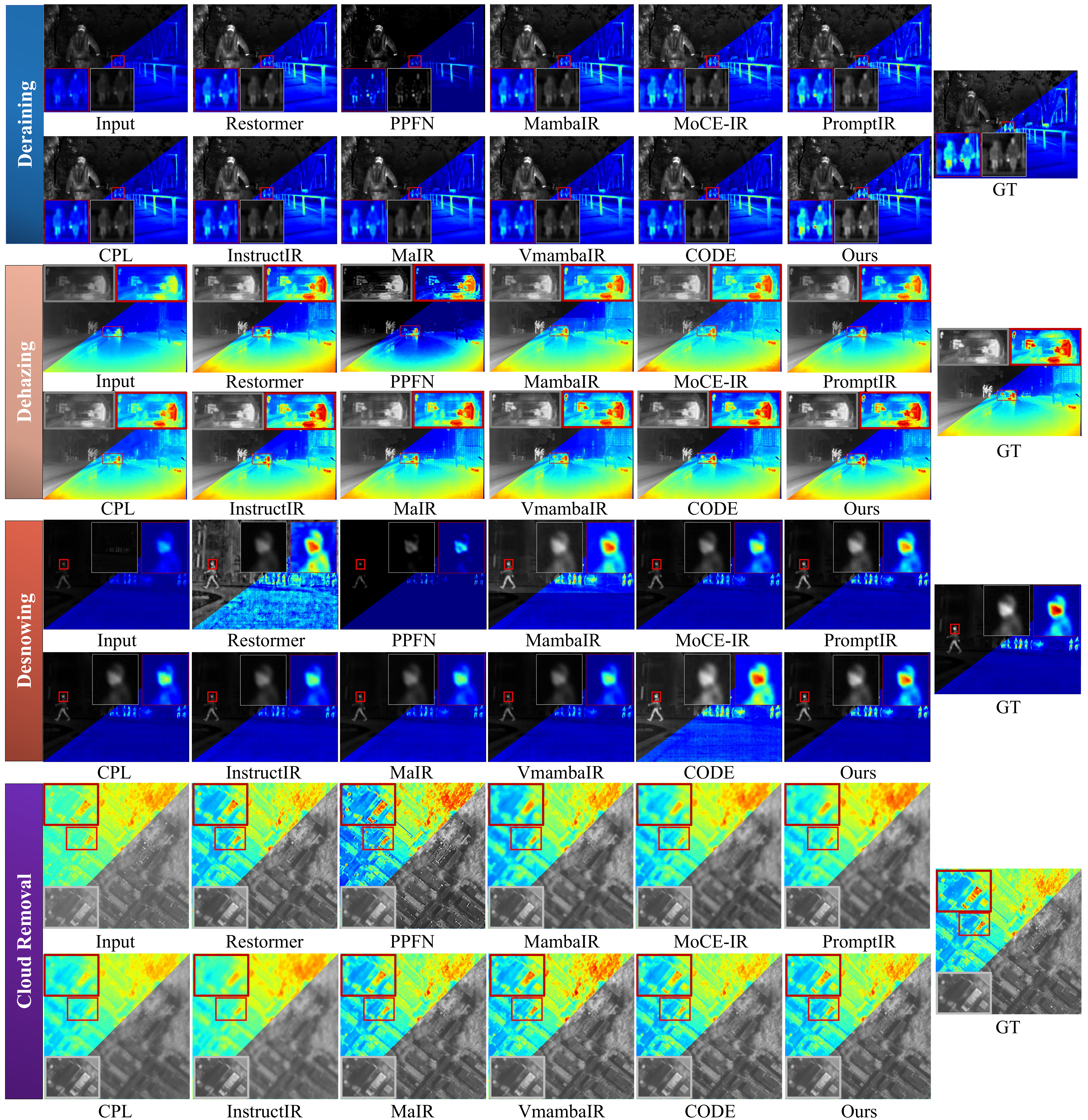}
\caption{Multi-weather degradation removal and thermal-structure fidelity comparison. Enlarged regions highlight weather-interference suppression, thermal-target preservation, and structural recovery under rain, haze, snow, and cloud degradations.}
\label{fig_5}
\end{figure*}

\noindent\textbf{Qualitative Analysis:}
Fig.~\ref{fig_5} presents visual comparisons under four degradation scenarios on the WeatherIR benchmark. Restormer introduces granular textures in some results. PPFN excessively suppresses background thermal responses, making certain targets and structures difficult to distinguish. In contrast, TSGPD-IR effectively removes different degradations while preserving scene structures and key thermal targets. For deraining, TSGPD-IR recovers clearer pedestrian contours and railing edges. In dehazing, it provides better contrast and thermal distribution recovery, resulting in clearer vehicles and backgrounds. For desnowing, it preserves limb structures and distant thermal targets. In cloud removal, it reconstructs more continuous road and building structures with fewer boundary discontinuities. Overall, TSGPD-IR achieves a better balance between degradation removal and structural preservation, producing results closer to the ground truth.

\subsubsection{Task-Specific Restoration}
\noindent\textbf{Quantitative Analysis.}
We further evaluate the effectiveness of TSGPD-IR under the task-specific setting, where an independent model is trained for each degradation. As shown in Table~\ref{tab:task_specific_comparison}, TSGPD-IR achieves the best PSNR and SSIM on all four tasks. It also obtains the lowest LPIPS for dehazing, desnowing, and cloud removal, while ranking second in LPIPS for deraining. For deraining, TSGPD‑IR achieves the best PSNR and SSIM, outperforming DRSformer by 0.0424 dB on PSNR, while its LPIPS is only slightly lower than DRSformer, indicating stronger pixel and structural fidelity with a minor gap on the perceptual metric. For dehazing, desnowing, and cloud removal, TSGPD-IR achieves the best overall performance across the three metrics. In particular, it improves PSNR by 2.0852~dB over BiLaLoRA for dehazing, by 0.3461~dB over SMGARN for desnowing, and by 1.1441~dB over CR-Former for cloud removal. These results confirm the effectiveness of TSGPD-IR when specialized to individual degradation types.

\begin{figure*}[pos=t]
\centering
\includegraphics[width=\linewidth]{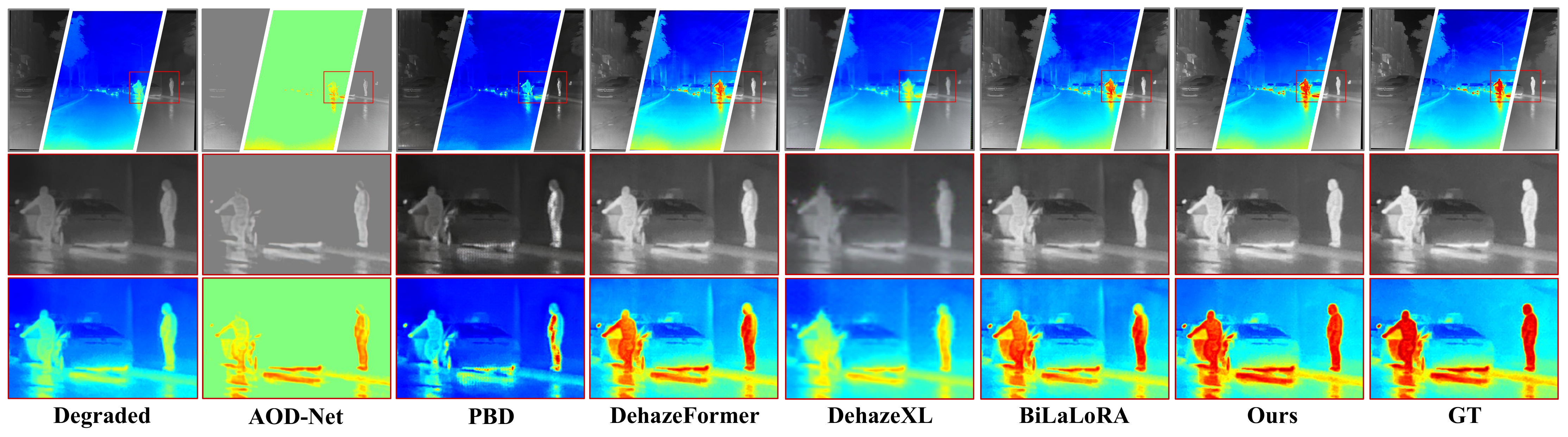}
\caption{Thermal-structure fidelity assessment under haze degradation. The top row shows the restored images, while the middle and bottom rows provide enlarged grayscale and pseudocolor views for examining thermal-response recovery and structural preservation.}
\label{fig_7}
\end{figure*}  
\begin{figure*}[pos=!t]
\centering
\includegraphics[width=\linewidth]{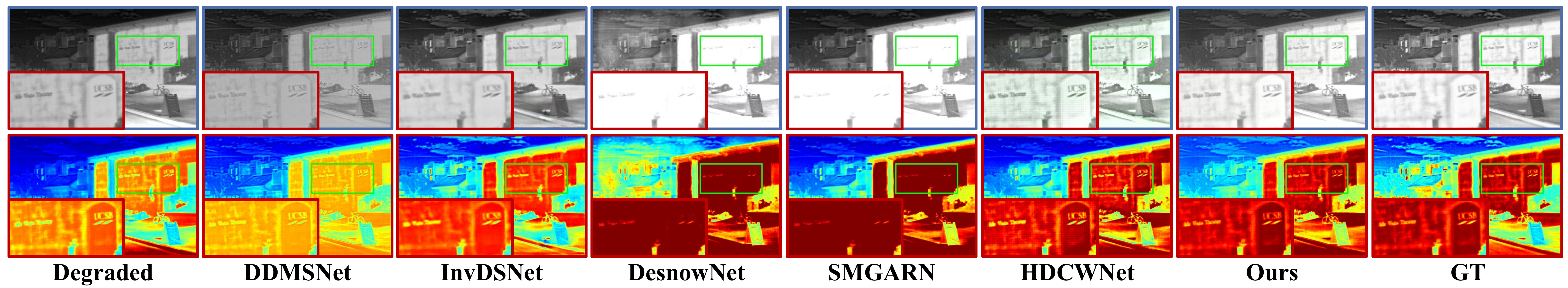}
\caption{Snow-interference suppression and thermal-structure fidelity assessment. The upper and lower rows show grayscale and pseudocolor results, respectively, while the enlarged regions highlight residual snow artifacts and the preservation of thermal structures.}
\label{fig_8}
\end{figure*}
\noindent\textbf{Qualitative Analysis.}
The qualitative comparisons further demonstrate the advantages of TSGPD-IR in preserving thermal responses and structural details across different degradation scenarios. 

\begin{figure*}[pos=t]
\centering
\includegraphics[width=\linewidth]{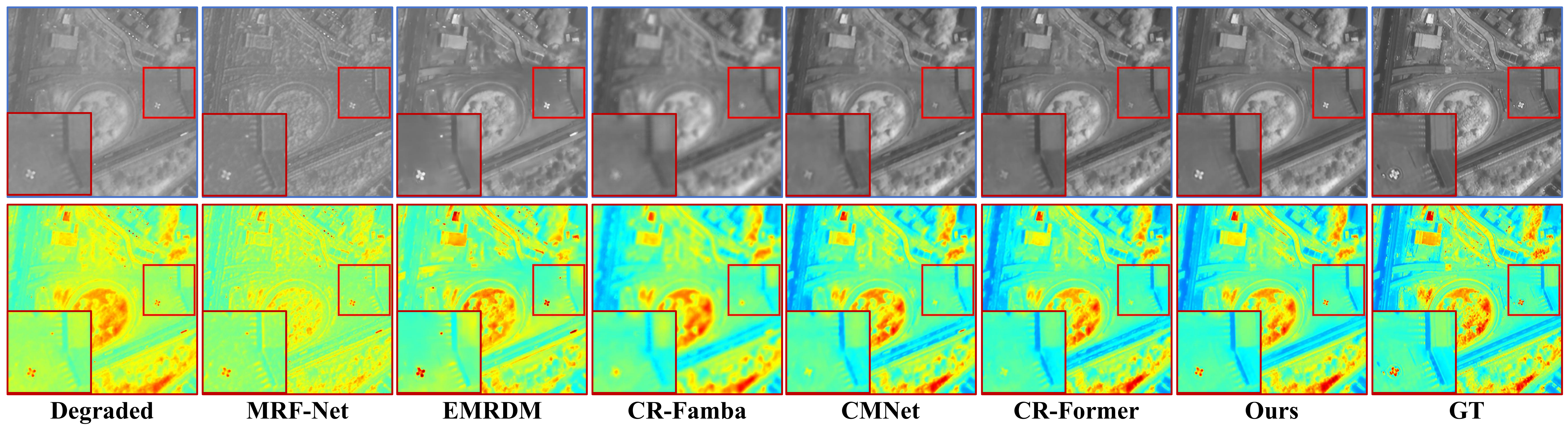}
\caption{Cloud-interference suppression and obscured-structure recovery assessment. The upper and lower rows show grayscale and pseudocolor results, respectively, while the enlarged regions highlight residual cloud contamination and the recovery of obscured road and building structures.}
\label{fig_9}
\end{figure*}

For dehazing, Fig.~\ref{fig_7} shows that AOD-Net causes severe intensity distortion, whereas PBD retains low contrast. Although DehazeFormer and BiLaLoRA improve target visibility, their results still exhibit local over-enhancement. In contrast, TSGPD-IR restores clearer pedestrian contours and foreground structures while maintaining a thermal distribution closer to the reference image. In the desnowing results of Fig.~\ref{fig_8}, DDMSNet weakens the thermal contrast of the building regions, whereas DesnowNet and SMGARN overenhance local thermal responses and lose structural details. TSGPD-IR better preserves building boundaries and textures, and recovers obscured background structures. For cloud removal, Fig.~\ref{fig_9} shows that MRF-Net and EMRDM retain cloud contamination. Although CMNet and CR-Former recover clearer scene content, local intensity inconsistencies remain. TSGPD-IR reconstructs more continuous road and building structures with fewer residual artifacts. Overall, these visual comparisons demonstrate that TSGPD-IR achieves a favorable balance among degradation removal, thermal-signal fidelity, and structural preservation.

\subsection{Ablation studies}
We conduct ablation studies on the WeatherIR benchmark to analyze the contribution of each component in TSGPD-IR under the four-task all-in-one setting.

\subsubsection{Effect of the Overall Framework of TSGPD-IR}
\begin{table}[pos=t]
\centering
\caption{Ablation study of the overall framework of TSGPD-IR on the WeatherIR benchmark. The best results are highlighted in \textbf{bold}.}
\label{tab:ablation_overall}
\setlength{\tabcolsep}{1.0pt}
\renewcommand{\arraystretch}{1.08}
\footnotesize
\resizebox{\columnwidth}{!}{\begin{tabular}{lcccc|ccc}
\toprule
Method & WS-MPG & PS-RDE & MCESS & Training &PSNR~$\uparrow$& SSIM~$\uparrow$ & LPIPS~$\downarrow$ \\
\midrule
baseline & $\times$ & $\times$ & $\times$ & $\times$ & 27.9405 & 0.8973 & 0.1182 \\
Weather-Semantic Pretraining & $\checkmark$ & $\times$ & $\times$ & I & 28.2931 & 0.9085 & 0.1076 \\
w/o MCESS & $\checkmark$ & $\checkmark$ & $\times$ & I$\rightarrow$II & 29.1297 & 0.9182 & 0.0955 \\ 	 	 
Weather-Pool Routing Only & $\checkmark$ & $\times$ & Type only & I$\rightarrow$II & 29.1167 & 0.9149 & 0.0965 \\
Severity-Expert Routing Only & $\times$ & $\checkmark$ & Severity only & I$\rightarrow$II & 29.2232 & 0.9181 & 0.1008 \\
One-Stage Joint Optimization& $\checkmark$ & $\checkmark$ & $\checkmark$ & Joint
& 26.3626 & 0.8787 & 0.1434 \\
\midrule
\rowcolor{red!10} \textbf{Ours}& $\checkmark$ & $\checkmark$ & $\checkmark$ & I$\rightarrow$II
& \textbf{29.5739} & \textbf{0.9213} & \textbf{0.0922} \\
\bottomrule
\end{tabular}}
\end{table}

Table~\ref{tab:ablation_overall} evaluates the contributions of the major components and the training strategy in TSGPD-IR. Starting from the baseline, weather-semantic pretraining improves the average PSNR from 27.9405~dB to 28.2931~dB and reduces LPIPS from 0.1182 to 0.1076, corresponding to an 8.97\% reduction, demonstrating the benefit of learning weather-aware restoration priors. Incorporating PS-RDE without MCESS further increases the PSNR to 29.1297 dB, while the complete model achieves 29.5739 dB, 0.9213 SSIM, and 0.0922 LPIPS, confirming that hierarchical expert routing effectively converts type and severity cues into adaptive restoration behaviors. Weather-pool routing and severity-expert routing alone obtain 29.1167 and 29.2232 dB, respectively, whereas their joint use yields gains of 0.4572 and 0.3507 dB, verifying their complementary roles in determining restoration direction and local processing strength. In contrast, one-stage joint optimization causes a substantial 3.2113 dB PSNR drop, validating the necessity of the proposed progressive two-stage training strategy.

\subsubsection{Effect of WS-MPG}
As shown in Table~\ref{tab:ablation_wsmpg}, we investigate the key design choices of WS-MPG. Removing prompt guidance or restricting prompt injection to Stage~I reduces the average PSNR by 0.4187~dB and 0.6449~dB, respectively, confirming the importance of prompt-conditioned modulation and progressive guidance. The Feature-Only Prompt and Weather-Only Prompt variants achieve average PSNR values of 29.1961~dB and 29.2030~dB, respectively, both below the complete model, demonstrating the complementarity between image-dependent features and explicit weather semantics. Removing the weather encoder causes a 0.3448~dB average PSNR drop, including a pronounced 0.9338~dB decrease on haze, while removing high-frequency cues reduces the average PSNR by 0.2571~dB, with larger degradation on rain and snow. These results indicate that weather-discriminative semantics and high-frequency information provide complementary cues for separating weather interference from intrinsic thermal structures.
\begin{table*}[pos=t]
\centering
\caption{Ablation study of WS-MPG on the WeatherIR benchmark.  The best and second-best results are highlighted in \textbf{bold} and \underline{underlined}, respectively.}
\label{tab:ablation_wsmpg}
\setlength{\tabcolsep}{3.2pt}
\renewcommand{\arraystretch}{1.10}
\scriptsize
\resizebox{\textwidth}{!}{\begin{tabular}{lccccccccccccccc}
\toprule
\multirow{2}{*}{Method} & \multicolumn{3}{c}{Rain} & \multicolumn{3}{c}{Haze} & \multicolumn{3}{c}{Snow} & \multicolumn{3}{c}{Cloud} & \multicolumn{3}{c}{Average} \\
\cmidrule(lr){2-4}
\cmidrule(lr){5-7}
\cmidrule(lr){8-10}
\cmidrule(lr){11-13}
\cmidrule(lr){14-16}
& PSNR~$\uparrow$ & SSIM~$\uparrow$ & LPIPS~$\downarrow$
& PSNR~$\uparrow$ & SSIM~$\uparrow$ & LPIPS~$\downarrow$
& PSNR~$\uparrow$ & SSIM~$\uparrow$ & LPIPS~$\downarrow$
& PSNR~$\uparrow$ & SSIM~$\uparrow$ & LPIPS~$\downarrow$
& PSNR~$\uparrow$ & SSIM~$\uparrow$ & LPIPS~$\downarrow$ \\
\midrule
w/o Prompts & 32.6072 & 0.9660 & 0.0363 & \underline{26.5118} & \underline{0.9593} & \underline{0.0504} & 31.8003 & 0.9514 & 0.0473 & 25.7016 & 0.7909 & 0.2521 & 29.1552 & 0.9169 & 0.0965 \\
Feature-Only Prompt & 32.8531 & 0.9675 & 0.0358 & 26.2200 & 0.9568 & 0.0517 & 31.9922 & 0.9530 & 0.0473 & 25.7192 & 0.7910 & 0.2489 & 29.1961 & 0.9171 & 0.0959 \\
Weather-Only Prompt & 32.8341 & 0.9668 & 0.0359 & 26.3619 & 0.9574 & 0.0520 & 31.9340 & 0.9519 & 0.0476 & 25.6818 & 0.7901 & \textbf{0.2485} & 29.2030 & 0.9166 & 0.0960 \\
Prompt Only at Stage~I & 32.2689 & 0.9574 & 0.0397 & 26.4318 & 0.9566 & 0.0521 & 31.3463 & 0.9409 & 0.0528 & 25.6689 & \underline{0.7915} & \underline{0.2488} & 28.9290  & 0.9116 & 0.0984 \\	w/o Weather Encoder & \underline{33.3503} & \textbf{0.9762} & \textbf{0.0283} & 25.6707 & 0.9516 & 0.0572 & \underline{32.2331} & \underline{0.9577} & 0.0409 & 25.6624 & 0.7911 & 0.2520 & 29.2291 & \underline{0.9192} & \underline{0.0946} \\
w/o High-Frequency & 33.0750 & 0.9700 & 0.0319 & 26.3598 & 0.9580 & 0.0513 & 32.0676 & 0.9528 & \underline{0.0457} & \textbf{25.7648} & 0.7929 & 0.2501 & \underline{29.3168} & 0.9184 & 0.0948 \\
\midrule
\rowcolor{red!10}Ours & \textbf{33.4397} & \underline{0.9714} & \underline{0.0311} & \textbf{26.6045} & \textbf{0.9625} & \textbf{0.0477} & \textbf{32.5130} & \textbf{0.9582} & \textbf{0.0409} & \underline{25.7383} & \textbf{0.7929} & 0.2490 & \textbf{29.5739} & \textbf{0.9213} & \textbf{0.0922} \\
\bottomrule
\end{tabular}}
\end{table*}
\begin{table*}[pos=t]
\centering
\caption{Ablation on proxy-supervision strategies in PS-RDE for the WeatherIR benchmark.
SROCC denotes the Spearman rank correlation between proxy labels and region-wise negative PSNR. Best results are highlighted in \textbf {bold}.}
\label{tab:ablation_psrde}
\setlength{\tabcolsep}{3pt}
\renewcommand{\arraystretch}{1.12}
\scriptsize
\begin{tabularx}{\textwidth}{l *{8}{>{\centering\arraybackslash}X}}
\toprule
\multirow{2}{*}{Setting} & \multicolumn{5}{c}{SROCC~$\uparrow$} & \multicolumn{3}{c}{Reconstruction Metrics} \\
\cmidrule(lr){2-6} \cmidrule(lr){7-9}
& Rain & Haze & Snow & Cloud & Avg. & PSNR~$\uparrow$ & SSIM~$\uparrow$ & LPIPS~$\downarrow$ \\
\midrule
w/o Proxy Supervision    & --  & --  & --  & --  & --  & 29.0000 & 0.9097 & 0.0988 \\
Residual-only Proxy      & 0.9769 & 0.9309 & 0.9762 & 0.9689 & 0.9632 & 29.3582 & 0.9182 & 0.0949 \\
Gradient-only Proxy      & 0.7760 & 0.7564 & 0.8304 & 0.3353 & 0.6745 & 29.3624 & 0.9190 & 0.0942 \\
\midrule
\rowcolor{red!10}Ours  & \textbf{0.9793} & \textbf{0.9339} & \textbf{0.9788} & \textbf{0.9930} & \textbf{0.9713} & \textbf{29.5739} & \textbf{0.9213} & \textbf{0.0922} \\
\bottomrule
\end{tabularx}
\vspace{-6pt}
\end{table*}
\begin{table}[pos=t]
\centering
\caption{Ablation study of the branch design and the number of severity-aware experts in MCESS on the WeatherIR benchmark. The best results are highlighted in \textbf{bold}.}
\label{tab:ablation_mcess}
\renewcommand{\arraystretch}{1.08}
\fontsize{7.5pt}{9pt}\selectfont

\resizebox{\columnwidth}{!}{
\begin{tabular}{lccc}
\toprule
Setting & PSNR~$\uparrow$ & SSIM~$\uparrow$ & LPIPS~$\downarrow$ \\
\midrule
w/o MCESS & 29.1297 & 0.9182 & 0.0955 \\
w/o Shared Branch & 29.2464 & 0.9199 & 0.0928 \\
w/o Adaptive Branch & 29.2769 & 0.9193 & 0.0955 \\
\midrule
$L=2$ & 29.5214 & 0.9199 & 0.0923 \\
$L=6$ & 29.5043 & 0.9205 & 0.0930 \\
\midrule
\rowcolor{red!10}
Ours & \textbf{29.5739} & \textbf{0.9213} & \textbf{0.0922} \\
\bottomrule
\end{tabular}
}
\end{table}

\subsubsection{Effect of Proxy Supervision in PS-RDE}
As shown in Table~\ref{tab:ablation_psrde}, we first assess the reliability of the constructed proxy labels using SROCC. The residual-only proxy achieves an average SROCC of 0.9632, showing that residual magnitude reliably reflects the overall intensity
deviation within each region. In contrast, the gradient-only proxy drops to 0.3353 on cloud, since gradient statistics are strongly affected by scene structures and cannot fully characterize the intensity attenuation and occlusion within cloud-covered regions. Combining the two cues increases the average SROCC to 0.9713 and achieves the best result for every weather type. The gain is particularly evident on cloud, where SROCC improves from 0.9689 to 0.9930, indicating that gradient variations complement residual magnitude in distinguishing structural perturbations from regional intensity deviations. Removing proxy supervision decreases the average PSNR by 0.5739~dB and SSIM by 1.26\%, while increasing LPIPS by 7.16\%. Residual-only and gradient-only supervision remain 0.2157~dB and 0.2115~dB below the complete model, respectively. This is because either cue alone yields incomplete severity estimates. Their joint use provides more reliable regional severity guidance and enables PS-RDE to regulate restoration strength accordingly.

\subsubsection{Effect of MCESS}
As shown in Table~\ref{tab:ablation_mcess}, we examine the dual-branch design and the number of severity-aware experts in MCESS. Removing MCESS decreases the average PSNR by 0.4442~dB and increases LPIPS by 3.58\%, confirming the contribution of expert-based adaptive restoration. Without the shared branch, the average PSNR drops by 0.3275~dB, as weather-specific experts must additionally model common thermal structures, thereby weakening their specialization in degradation removal. Removing the adaptive branch leads to a 0.2970~dB PSNR reduction and a 3.58\% increase in LPIPS, since the shared pathway alone cannot allocate restoration capacity according to weather type and regional severity. These results demonstrate that the shared branch provides structure-preserving representations, while the adaptive branch performs degradation-specific correction. We further vary the number of experts in each weather-specific pool. Compared with the four-expert configuration, using two or six experts reduces the average PSNR by 0.0525~dB and 0.0696~dB, respectively. Two experts provide insufficient granularity for severity-specific restoration, whereas six experts may fragment the routed samples and weaken expert optimization. Therefore, four experts achieve a better balance between expert specialization and routing stability.

\section{Conclusion}
We propose TSGPD-IR, a type-severity guided progressive disentanglement framework for all-in-one adverse-weather thermal infrared image restoration. Unlike existing infrared restoration methods that mainly focus on a single degradation type or directly transfer visible-image restoration paradigms, TSGPD-IR aims to disentangle weather-induced degradation from intrinsic thermal structures while preserving weak target contours and meaningful thermal details. The framework is built upon WS-MPG, PS-RDE, and MCESS. This design injects weather-semantic prompt guidance into stage-wise restoration features, estimates spatially varying degradation severity without manual region-level annotations, and enables hierarchical expert routing from weather-specific task pools to severity-compatible experts.

Experiments in both all-in-one and task-specific settings show that TSGPD-IR maintains high structural fidelity and perceptual quality and consistently improves restoration results for rain, haze, snow, and cloud degradations. These results suggest that all-in-one infrared image restoration need not rely solely on degradation-specific modeling or uniform restoration strategies. Instead, jointly modeling global weather type and regional degradation severity provides a more adaptive and controllable direction for suppressing diverse weather artifacts while preserving intrinsic thermal structures.

\printcredits
\section*{Declaration of Competing Interest}
The authors declare that they have no known competing financial interests or personal relationships that could have appeared to influence the work reported in this paper.

\section*{Acknowledgments}
This work is supported by the National Natural Science Foundation of China under Grant No. 62531012, the Sichuan Science and Technology Program under Grant 2026YFHZ0205, and the XJTU Research Fund for AI Science, No.2025YXYC004. 

\section*{Data availability}
Data will be made available on request.

\bibliographystyle{unsrtnat}

\bibliography{cas-refs}

\vspace*{-15pt}
\bio{figs/wxy}
{Xinyao Wang} received the B.S. degree from the College of Communication Engineering, Jilin University, Changchun, China, in 2026. She is currently pursuing her M.S. degree in the School of Information and Communications Engineering, Xi'an Jiaotong University. Her research interests include infrared image enhancement.
\endbio
\bio{figs/helijun}
{Lijun He} received the B.S. and Ph.D. degrees from the School of Information and Communications Engineering, Xi’an Jiaotong University, Xi’an, China, in 2008 and 2016, respectively. She is currently a Professor with the School of Information and Communications Engineering, Xi’an Jiaotong University. Her research interests include video communication and transmission, video analysis, processing, and compression techniques.
\endbio
\bio{figs/rzh}
{Zhihan Ren} received the B.S. degree from the College of Communication Engineering, Jilin University, Changchun, China, in 2023. He is currently pursuing his Ph.D degree with the School of Information and Communications Engineering, Xi'an Jiaotong University. His research interests include generative models, image super-resolution, and AI safety.
\endbio
\bio{figs/lifan}
{Fan Li} received the B.S. and Ph.D. degrees in information and communication engineering from Xi'an Jiaotong University, Xi'an, China, in 2003 and 2010, respectively.
From 2017 to 2018, he was a Visiting Scholar with the Department of Electrical and Computer Engineering, University of California at San Diego, San Diego, CA, USA. He is currently a Professor with the School of Information and Communications Engineering, Xi'an Jiaotong University. He has published more than 80 technical articles.
His research interests include multimedia communication, image/video coding, and image/video quality assessment.
\endbio

\end{document}